%% file: eacl2023.tex
\pdfoutput=1

\documentclass[11pt]{article}

\usepackage[]{EACL2023}

\usepackage{times}
\usepackage{latexsym}
\usepackage{amsfonts}
\usepackage[T1]{fontenc}

\usepackage[utf8]{inputenc}

\usepackage{microtype}

\usepackage{inconsolata}

\usepackage{booktabs} 
\usepackage{multirow}

\usepackage{array}
\usepackage{makecell}

\usepackage{cleveref}
\usepackage{subcaption}
\usepackage{graphicx}

\title{Revisiting Offline Compression: Going Beyond Factorization-based Methods for Transformer Language Models}

\author{Mohammadreza Banaei$^{*1}$, Klaudia Bałazy$^{*2}$, Artur Kasymov$^2$ \\ {\bf Rémi Lebret$^1$, Jacek Tabor$^2$, Karl Aberer$^1$}\\ $^1$EPFL, $^2$Jagiellonian University \\
}

\begin{document}

\maketitle

\def\thefootnote{*}\footnotetext{Equal contribution}\def\thefootnote{\arabic{footnote}}
\def\thefootnote{}\footnotetext{Correspondence to: \nolinkurl{mohammadreza.banaei@epfl.ch, klaudia.balazy@doctoral.uj.edu.pl}}\def\thefootnote{\arabic{footnote}}

\input{abstract}

\input{introduction}

\input{related_works}

\input{model}
\input{ablation_study}
\input{conclusions}

\input{limitations}
\input{ethic_statement}
\input{acknowledgement}







\bibliography{eacl2023}
\bibliographystyle{acl_natbib}


\input{apendix}


\end{document}

%% file: abstract.tex
\begin{abstract}
Recent transformer language models achieve outstanding results in many natural language processing (NLP) tasks. However, their enormous size often makes them impractical on memory-constrained devices, requiring practitioners to compress them to smaller networks.
In this paper, we explore offline compression methods, meaning computationally-cheap approaches that do not require further fine-tuning of the compressed model. 
We challenge the classical matrix factorization methods by proposing a novel, better-performing autoencoder-based framework. We perform a comprehensive ablation study of our approach, examining its different aspects over a diverse set of evaluation settings. Moreover, we show that enabling collaboration between modules across layers by compressing certain modules together positively impacts the final model performance. 
Experiments on various NLP tasks demonstrate that our approach significantly outperforms commonly used factorization-based offline compression methods.\footnote{Our code is public:~\href{https://github.com/MohammadrezaBanaei/auto-encoder-based-transformer-compression}{github.com/MohammadrezaBanaei/auto-encoder-based-transformer-compression}}.


\end{abstract}




%% file: introduction.tex
\section{Introduction}
\label{sec:intro}
The recent trend of pre-training Transformer~\cite{vaswani2017attention} language models on enormous unsupervised corpus has led to outstanding performances on many downstream tasks. For downstream tasks, these pre-trained models are then either fine-tuned~\cite{devlin-etal-2019-bert,liu2019roberta,yang2019xlnet} or the \emph{prompting} paradigm~\cite{brown2020language} is used (especially in the so-called Large Language Models), which avoids having a different model per task~\cite{lieber2021jurassic, rae2021scaling,smith2022using,thoppilan2022lamda}. In each of the two paradigms, it has been shown that increasing the scale of language models generally leads to better performance on a range of downstream tasks~\cite{devlin-etal-2019-bert,brown2020language}. Indeed, for autoregressive language models, \citet{kaplan2020scaling} demonstrated a power-law relationship between the number of parameters and the respective performance. \citet{wei2022emergent} further showed that certain abilities of language models emerge only when the number of its parameters passes certain thresholds, providing an incentive to scale these models further.

Although scaling up these language models make them empirically powerful across many diverse tasks, it makes them infeasible to train for many NLP practitioners due to huge pre-training costs. More importantly, even using the available pre-trained models for inference is becoming more challenging (especially for memory-constrained applications like edge devices), with recent models having hundreds of billions of parameters~\cite{zhang2022opt}.

With the rise of NLP model sizes, there have been many efforts to compress transformer language models without compromising their performance. Although being inherently different, many of these efforts rely on knowledge distillation~\cite{hinton2015distilling} to help the compressed model (i.e., the student model) better imitate the parent model (i.e., the teacher model). However, these approaches often need costly distillation on upstream or downstream tasks~\cite{sanh2019distilbert} as well as expensive data augmentation techniques~\cite{jiao2019tinybert} to help improve the compressed model performance. These approaches become even less feasible when enormous language models are being distilled.

Another line of research focuses on computationally-cheap methods (i.e., offline compression) where a smaller model can be achieved from a pre-trained model without it being necessarily fine-tuned over a downstream or upstream task. These offline methods include weight pruning~\cite{li2016pruning, han2015learning}, weight quantization~\cite{zhou2016dorefa, hubara2016binarized}, tensor factorization~\cite{lan2019albert, winata2019effectiveness, balazy-etal-2021-direction, cordonnier2020multi} and hybrid approaches~\cite{wang2019structured, mao2020ladabert}. 

This paper proposes a novel offline factorization-based method for compressing transformer language models. 
The paper's main goal is to propose an offline method that produces a competitive language model (compared to the original model perplexity) before any fine-tuning is performed. 
Similar to~\citet{balazy-etal-2021-direction}, we use an autoencoder model (see~\Cref{fig:aemainfig}) to compress different modules' weights. However, unlike the previous work, our approach is not limited to the token embeddings (a.k.a. word embeddings) and can be applied to other transformer modules as well. We also propose and thoroughly investigate the impact of various enhancements for the approach of obtaining the compressed model. It is worth noting that although the experiments and ablation studies are only done BERT$_{BASE}$ model, our 

In~\Cref{sec:ablation} we demonstrate that applying small changes to the autoencoder architecture (e.g., introducing non-linearity to the decoder) and its loss objective results in superior performance to the Singular Value Decomposition (SVD) baseline as measured by model perplexity and its performance on the downstream tasks.
Moreover, inspired by the redundancies present across self-attention heads~\cite{cordonnier2020multi}, in~\Cref{sec:abl:concat_separate} we show that the compressed models perform in general better when compressing certain modules from different layers together.

Additionally, in~\Cref{sec:abl:sensitivity} we investigate the effectiveness of a (parameter) sensitivity-based\footnote{We call it sensitivity as it measures how sensitive the model performance is to the reconstruction error of a certain parameter.} compression by incorporating fisher information~\cite{pascanu2013revisiting} in the loss objective. We later show that incorporating these weights significantly improve the compressed language model performance (i.e., perplexity).

Finally, in~\Cref{sec:abl:baselines} and in~\Cref{sec:abl:finalres} we discuss the performance of our approach in comparison to various offline-compression baselines and demonstrate that our method provides the best or competitive quality of the compressed model.


Our main contribution can be summarized as follows:

\begin{itemize}
    \item We propose a novel autoencoder-based framework for low-cost compression of transformer language models and conduct an extensive ablation study on its different aspects. 
    \item We show that enabling collaboration across layers by compressing different layers modules together boosts the performance.
    \item We demonstrate that our approach significantly outperforms other commonly used offline-compression methods on various NLP downstream tasks.\footnote{It is worth noting that although the experiments and ablation studies in this paper are only done on the BERT$_{BASE}$ model, our proposed approach can potentially be used for any transformer-based architecture.}
 \end{itemize}



%% file: related_works.tex
\section{Related work}
\label{sec:related_work}
Deep transformer language models have gained increasing attention in recent years since the seminal work of~\citet{devlin-etal-2019-bert}. Many recent efforts demonstrate that scaling up these language models' parameters generally results in better performance on a range of downstream tasks~\cite{devlin-etal-2019-bert, brown2020language,kaplan2020scaling}. This empirical observation resulted in recent language models having over a thousand times more parameters~\cite{lieber2021jurassic,rae2021scaling,smith2022using} than the BERT$_{BASE}$ model~\cite{devlin-etal-2019-bert}. Although empirically powerful, these models are becoming harder to use for memory-constrained applications, which led to many efforts toward language model compression in recent literature.

Although many recent efforts for language model compression take advantage of distillation~\cite{hinton2015distilling} techniques to better imitate the uncompressed model (i.e., the teacher model) behavior, this paper focuses mainly on \emph{offline} compression methods. By \emph{offline}, we refer to approaches that do not need fine-tuning the whole model on a downstream/upstream dataset. In the case of language model compression, these methods aim to output a compressed model without losing too much performance (measured by perplexity) that can then be fine-tuned (with or without distillation) or prompted for a certain downstream task. It is worth noting that these methods can still be combined with distillation techniques, but starting the finetuning from a better compressed language model would generally reduce its costs of training (e.g., by improving convergence time).

Offline compression methods, while being diverse, can be roughly categorized into few paradigms, namely weight pruning~\cite{see2016compression, li2016pruning, han2015learning, fan2019reducing, michel2019sixteen, voita2019analyzing}, quantization~\cite{gong2014compressing, hubara2016binarized, zhou2016dorefa}, tensor factorization~\cite{lan2019albert, winata2019effectiveness, balazy-etal-2021-direction, cordonnier2020multi,shapeshifter,ren2022exploring} and hybrid approaches~\cite{wang2019structured, mao2020ladabert}.

This paper primarily focuses on the effectiveness of low-rank factorization-based approaches in recent literature~\cite{lan2019albert,shapeshifter,ren2022exploring} as an \emph{offline} compression method. \citet{lan2019albert} proposed a SVD-based~\cite{halko2011finding} technique to compress the token embedding module. Later works have shown that more complex architectures like autoencoders can result in better compression quality than SVD methods~\cite{lioutas2020improving,balazy-etal-2021-direction}. By taking advantage of the autoencoder, we are able to more easily enforce different properties by either changing its training objective or architecture (e.g., preserving $l_2$ norm in reconstructed embeddings). \citet{balazy-etal-2021-direction} emphasized on the importance of \emph{direction} in token embedding compression, and in this work we demonstrate its potential importance for other transformer modules as well in different compression ratios.

Moreover, \citet{cordonnier2020multi} showed the significance of redundant information in self-attention heads and compressed different heads (in a certain layer) together to improve compression performance. Following a similar idea, we later show that compressing heads from different layers together would generally further boost the compression quality.

\citet{hsu2022language} proposed a weighted SVD (using Fisher Information)~\cite{pascanu2013revisiting} to outperform the classical SVD. We further investigate the benefits of a non-uniform compression (i.e., a weighted reconstruction loss in the autoencoder loss objective) in~\Cref{sec:abl:sensitivity} by analyzing different weighting schemes for parameters.

Moreover, \citet{ren2022exploring} proposed using tensor decomposition techniques to compress language models to relatively high compression ratios while using a two-stage distillation technique. Moreover, \citet{shapeshifter} proposes using the Kronecker product as an alternative for the factorization of transformer modules. \Cref{sec:abl:kronecker_tucker} discusses using Tucker~\cite{de2000multilinear} or Kronecker-based methods as an offline approach. It is worth noting that models with relatively high compression ratios become highly dependent on distillation techniques to perform reasonably on downstream tasks. For instance, \citet{ren2022exploring} claims that even randomly initializing the compressed BERT nearly achieves identical performance compared to tensor decomposition from a pre-trained model.

%% file: model.tex
\section{Model}
\label{sec:model}

Our offline compression approach is based on the an autoencoder neural network architecture, similar to~\citet{lioutas2020improving} and~\citet{balazy-etal-2021-direction}. However, in this work, we focus on compressing all the transformer weight matrices rather than just the token embedding matrix. Furthermore, we are exploring many more compression improvements using autoencoder as well as investigating architecture-independent techniques.

\begin{figure}[ht]
    \centering
    \begin{subfigure}[b]{\columnwidth}
    \centering
    \includegraphics[width=1.0\columnwidth,height=.36\columnwidth]{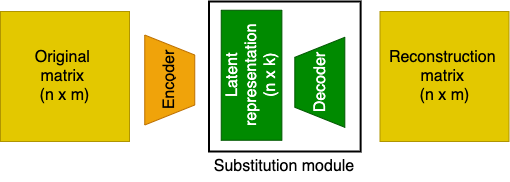}
    \caption{Autoencoder-based compression with
customizable objective function. Our approach minimizes the root mean square error (RMSE) and cosine distance between the original and reconstruction matrix. In this setting, the original matrix's latent representation and the decoder form the substitution module.\label{fig:aemainfig}}
    \end{subfigure}
    \vskip 0.1in
    \begin{subfigure}[b]{\columnwidth}
    \includegraphics[width=1.0\columnwidth,height=.36\columnwidth]{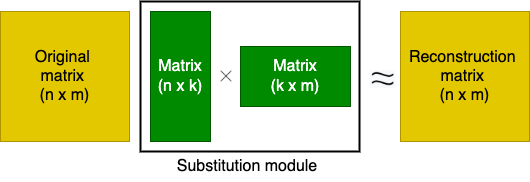}
    \caption{Classical matrix-factorisation-based compression with root mean square error (RMSE) minimization objective. Two smaller matrices, approximating the target matrix after multiplication, form the substitution module.\label{fig:svdmainfig}}
    \end{subfigure}
    \caption{A high-level view of the matrix compression approaches using classical matrix-factorization and autoencoder model that is leveraged in this work. The purpose of the compression is to provide a parameter-efficient substitution module to replace the original matrix in the considered model.\label{fig:mainfig}}
\end{figure}

The autoencoder architecture consists of encoder function $h(\cdot)$ that maps model input $x\in\mathbb{R}^m$ to some latent representation $l\in\mathbb{R}^k$. The second element of the architecture is the decoder function $g(\cdot)$ responsible for mapping  $l\in\mathbb{R}^k$ into the approximation~$\tilde{x}\in\mathbb{R}^m$ of the input $x$. 

Let us assume that we want to compress the matrix $A\in\mathbb{R}^{(n \times m)}$. Using the gradient-descent algorithm, we train the autoencoder model to produce the appropriate approximation $\tilde{A}$ of the original matrix. As a compressed module, we understand the hidden representation $h_{\Psi}(A) \in \mathbb{R}^{(n \times k)}$ together with the decoding module $g_{\Phi}(\cdot)$. In this setting, the formula for the \emph{compression ratio} of the original module can be expressed as:
\begin{equation}
    \frac{n \cdot m}{(n \cdot k) + |\Phi|},
\end{equation}
which is the ratio of the original matrix size to the hidden representation size and the number of parameters in the decoder module~$|\Phi|$. We illustrated the approach of compressing a matrix using the autoencoder model in~\Cref{fig:aemainfig}.

Our approach to offline compression based on the autoencoder offers flexibility in performing the ablation study as we are able to easily modify its elements, for example, decoder module complexity level or the loss function components. By using an autoencoder architecture with a linear decoder and the RMSE cost function, we can obtain the equivalent approximation as provided by a simple matrix factorization. An illustrative comparison of our compression method with the classical matrix factorization approach is shown in~\Cref{fig:mainfig}.

Following~\citet{balazy-etal-2021-direction}, we train our autoencoder with the multi-objective cost function consisting of $l_2$~norm loss and cosine distance loss:
\begin{equation}
    \label{eq:l2cosloss}
    \Psi_{\beta}(X, \widetilde{X})  = (1-\beta) \cdot L_2(X, \widetilde{X}) + \beta \cdot CD(X, \widetilde{X}),
\end{equation}
where $X$ represents the original matrix, $\widetilde{X}$ is the reconstructed matrix, $L_2(X, \widetilde{X})$ represents the root mean square error (RMSE) loss function, and $CD(X, \widetilde{X})$ is the mean cosine distance loss for all pairs of vectors (rows) of the original and reconstructed matrices. The $\beta$ hyperparameter ($0 \leq \beta \leq 1$) is responsible for determining the weight we would like to assign to the different components of the loss function.


%% file: ablation_study.tex
\section{Experiments}
\label{sec:ablation}

This section describes our motivations and the results of various analyses and experiments that we conducted to investigate the topic of offline compression thoroughly.

We performed our experiments for different weight matrices in the transformer architecture, as each type of weight matrix may have different characteristics, and a given compression method may or may not be appropriate. The analyses described below are performed for token embedding, self-attention (keys, queries, values), and output-dense weight matrices.

We focus our study on the BERT$_{BASE}$ model~\cite{devlin-etal-2019-bert}, but the same methods could be applied to other transformer architectures. All experiments are conducted for three compression ratios (3, 10, and 25) to investigate the differences given the different number of available parameters. All experimental settings of the various studies presented in the following sections are included in~\Cref{sec:appendix1}.

We evaluate the quality of our compressed models on the masked~\cite{devlin-etal-2019-bert} language modeling task (using the WikiText-103 test dataset~\cite{merity2016pointer}) and multiple datasets from the GLUE benchmark~\cite{wang-etal-2018-glue}.


\subsection{Cosine distance objective}
\label{sec:abl:cos_dis_obj}


First, we investigate whether including the direction component in the compression objective has a positive effect on the compression of weight matrices other than token embeddings in the transformer model. \citet{balazy-etal-2021-direction} demonstrated that supplementing the loss function with the cosine distance between pairs of rows of the original and reconstructed matrix produces noticeably better compression results for the token embeddings matrix. Unfortunately, their study does not examine other matrices in the transformer, whereas because of the different nature of these matrices, we believe it is worth investigating.

\paragraph{Results}
In \Cref{tab:coskey} and \Cref{tab:cosout} (in \Cref{sec:appendix1:cos}), we present the effect of adding the cosine distance component to the cost function for the keys and output-dense matrices from the BERT$_{BASE}$ model. It seems that for matrices other than token embeddings, considering the direction of vectors (rows in the matrix) in most cases may have a positive impact on the final results of the compressed model. However, the benefits of using this component are not as significant as in the case of the token embeddings matrix. Indeed, there are some examples where minimizing only Euclidean distance or adding only a small proportion of cosine distance provides the best results. We suppose this behavior is the consequence of the token embeddings matrix nature, where the rows represent specific tokens used to construct the words. It seems that the representation of the part or entire word is largely encoded in the vector direction. This characterization does not necessarily apply to other matrices though often there is a subtle benefit from adding a cosine distance component to the reconstruction objective.

\subsection{Concatenated and separated weight matrices}
\label{sec:abl:concat_separate}
\begin{figure}[ht]
    \centering
    \begin{subfigure}[b]{\columnwidth}
    \centering
    \includegraphics[width=0.7\columnwidth,height=.7\columnwidth]{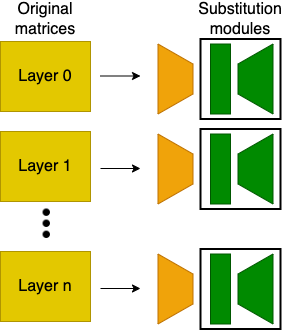}
    \caption{Separated matrices compression mode. All substitution modules have separate decoder. \label{fig:sep}}
    \end{subfigure}
    \vskip 0.1in
    \begin{subfigure}[b]{\columnwidth}
    \centering
    \includegraphics[width=.7\columnwidth,height=.6\columnwidth]{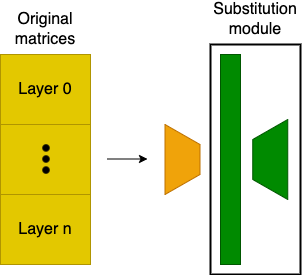}
    \caption{Concatenated matrices compression mode. All substitution modules share the decoder module that enables collaboration between layers. \label{fig:conc}}
    \end{subfigure}
    \caption{Separated and concatenated weight matrices compression. We demonstrate that compressing concatenated matrices from all layers provides a better-performing substitution module. In the concatenated setting, the substitution modules share a decoder that allows for cross-layer collaboration and provides the potential to further eliminate redundant information.\label{fig:sepvsconcat}}
\end{figure}

\begin{figure*}[ht]
    \centering
    \includegraphics[scale=0.33]{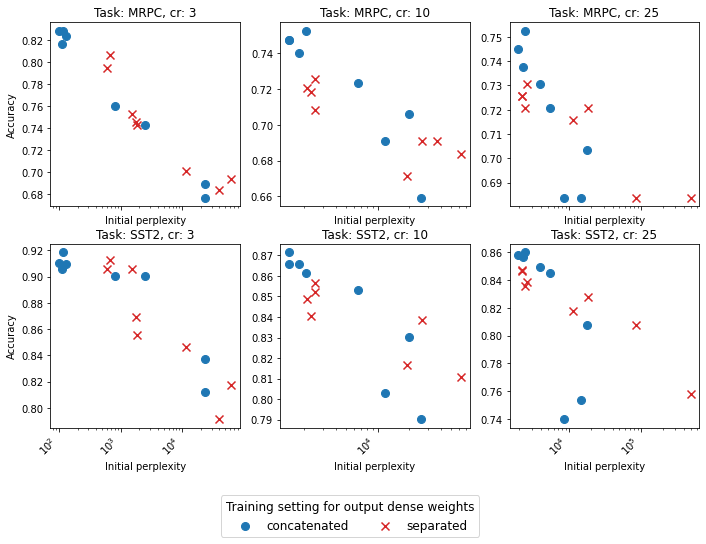}
    \caption{\label{fig:concvssep1}Initial perplexity and downstream tasks performance of output-dense matrices' compression in separated/concatenated compression modes\footnotemark. Using the concatenated mode generally results in better performance for this module. We observe similar pattern for key, query, and value weight matrices~(see~\Cref{fig:concvssep2} in~\Cref{sec:appendix1:concsep}).}

    
\end{figure*}

This section investigates whether compressing separately each weight matrix from the considered model is the best possible strategy. Our intuition is that compressing together the same type of weight matrices from different layers may bring certain advantages. First, it could allow for minimizing redundant information in the model weights, and second, it could enable collaboration between compressed modules across different layers.
Suppose that the neural network model consists of $n$ layers $(l_0, l_1, \dots, l_{n-1})$. Each layer $l_i$ encapsulates a particular weight matrix $W_{l_i}$. Conventionally, each $W_{l_i}$ matrix is considered separately during the compression process. In the concatenated mode, we propose compressing a single matrix $W=[W_{l_0}, W_{l_1}, \dots, W_{l_{n-1}}]$ resulted from concatenating all $W_{l_i}$ matrices. Given the proposed compression process, the compressed weight matrices share a common decoder as illustrated in~\Cref{fig:sepvsconcat}.

\paragraph{Results}

Experiments discussed in this section demonstrate that compressing concatenated weight matrices performs better than compressing each matrix separately in terms of the compressed model performance as well as the compression process time. \Cref{fig:concvssep1} presents the performance achieved by models with compressed output-dense matrices in separated and concatenated modes (similar experiments are presented for key, query, and value matrices in~\Cref{fig:concvssep2} in~\Cref{sec:appendix1:concsep}). We report the initial perplexity and the final score achieved on the MRPC and SST2 downstream tasks for different compression ratios. We observe the apparent dominance of the concatenated mode over the separated mode for both perplexity and downstream task performance. This may indicate that sharing the decoder helps to reduce redundant information and saves parameters for further knowledge encoding.

Furthermore, the separated compression mode is more computationally expensive at the initial stage as we must compress each matrix individually. In the concatenated mode, we perform only a single compression process on the matrices' concatenation. 

Considering the better performance and faster training process, we only analyze the concatenated weight matrices compression in the following sections.

\subsection{Initial perplexity vs downstream tasks performance}
\label{sec:abl:perplacc}


Perplexity is a popular measure determining how well the language model predicts a particular sequence of tokens (the lower the perplexity the better). This section introduces that the masked language model perplexity metric may be considered as a low-cost yet effective way to evaluate a compressed module. We show that in most cases the compressed models with the lowest initial perplexity yields the best performance when fine-tuned on a downstream tasks.

\footnotetext{Each point represents one hyperparameter setting.}

\paragraph{Results}
We examined the relation between the initial BERT$_{BASE}$ perplexity after applying compressed weight matrices and its final performance after fine-tuning on a downstream task. In~\Cref{fig:perplvsacc2} (in~\Cref{sec:appendix1:initperp}) we report results for token embeddings matrix, self-attention keys, queries, values and final output-dense matrices. We observe that in most cases models with the lowest initial perplexity result in the best performance on the downstream task (MRPC ans SST2). Therefore, we consider the masked language model perplexity metric to be a good low-cost method to preliminarily evaluate the quality of a compressed module. 



\subsection{Linear and non-linear decoder module}
\label{sec:abl:linnonlin}
In this section we investigate the effect of using different decoder module in the autoencoder model on the final model's performance. We experiment with a simple linear layer decoder and two non-linear decoder versions.

\begin{figure*}[t]
    \centering
    \includegraphics[width=1.25\columnwidth]{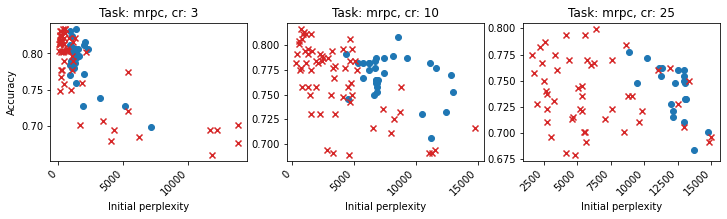}
    \includegraphics[width=1.25\columnwidth]{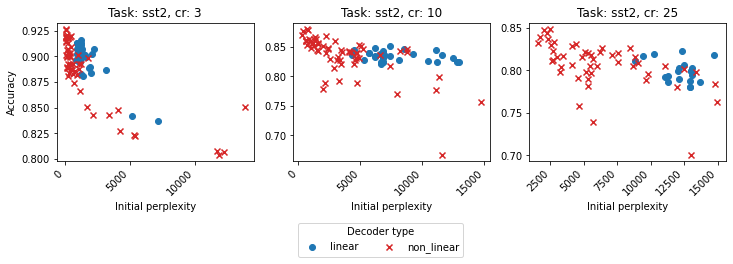}
    \caption{Initial perplexity and downstream tasks performance for the compressed token embeddings matrix when using either a linear or a non-linear decoder module in the autoencoder model. Here we present results for different hyperparameters settings (see~\Cref{sec:appendix_linear} for further details). A non-linear decoder seems to be a better choice for compressing this module.}
    \label{fig:linvsnonlin-te}
\end{figure*}


\paragraph{Results}
\Cref{fig:linvsnonlin-te} presents initial perplexity and final downstream tasks performance achieved when using linear and non-linear decoder in the autoencoder model while compressing token embeddings matrix. We may observe that for the token embeddings better final results are produced when using non-linear decoder.  However, as demonstrated in~\Cref{fig:lin_non_lin_key} (in~\Cref{sec:appendix_linear}), a different pattern is apparent for key matrices where linear models considerably outperform the non-linear versions in most cases.

\subsection{Preserving vector norm}

Furthermore, we examine whether preserving the original $l_2$ vector norms of the vectors representing rows in the reconstructed matrix to be the same as in the original vectors is beneficial for the compression.



\begin{figure*}[t]
    \centering
    \includegraphics[width=1.25\columnwidth]{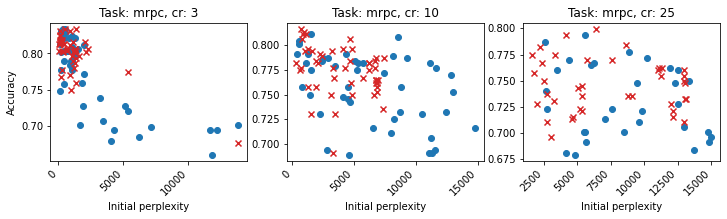}
    \includegraphics[width=1.25\columnwidth]{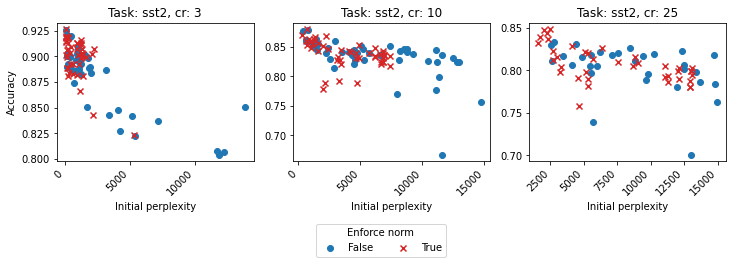}
    \caption{The effect of preserving $l_2$ vector norm on the perplexity and downstream tasks performance while compressing the token embeddings matrix for different hyperparameters settings. Preserving norm seems to generally improve the compression performance for this module.}
    \label{fig:enforcenorm-te}
\end{figure*}

\paragraph{Results}
\Cref{fig:enforcenorm-te} presents initial perplexity and downstream tasks performance when enabling or disabling the preserving vector norm technique for the token embeddings matrix. We may see that in most cases the version with enabled preserving vector norm achieves better results. 
In addition to token embeddings, in~\Cref{fig:enforce_norm_od} (in~\Cref{sec:app:preservingnorm}) we also demonstrate the effect of preserving $l_2$ vector norm during compression of the output-dense matrix.

\subsection{Sensitivity}
\label{sec:abl:sensitivity}

Most offline compression methods focus only on the raw weight matrices taken from the considered pre-trained model. However, we could also leverage the unsupervised upstream dataset to improve the compression quality. \citet{hsu2022language} proposed using additional weights computed on the entire upstream dataset to enhance the low-rank factorization method. They used the Fisher Information weights~$I$ that measure the amount of the observable information in dataset $D$ about a single model parameter $w$. A feasible approximation~$\hat{I}_w$ of the Fisher Information $I_w$ for parameter $w$ may be expressed as: 
\begin{equation}
    \label{eq:fisher_emp}
    \resizebox{0.895\hsize}{!}{$I_w = E[ (\frac{\partial}{\partial w}\log P(D|w))^2] \approx\frac{1}{|D|} \sum_{d\in D}(\frac{\partial}{\partial w}L(d;w))^2=\hat{I}_w$}
\end{equation}
where $L$ is the target pre-training task objective (e.g., cross-entropy or MSE).

For the entire weights matrix $W$, \citet{hsu2022language} presented even more simplified and computationally effective row-wise diagonal Fisher Information matrix~$\hat{I}$, where each diagonal value is the sum of the corresponding row of the Fisher Information approximation matrix $\hat{I}_W$:

\begin{equation}
\label{eq:fisher_diag}
\hat{I} = diag(\sqrt{\sum_{j=1}\hat{I}_{W_{1j}}}, ..., \sqrt{\sum_{j=1}\hat{I}_{W_{nj}}}).
\end{equation}

\begin{figure}[h]
    \centering
    \includegraphics[width=1.0\columnwidth]{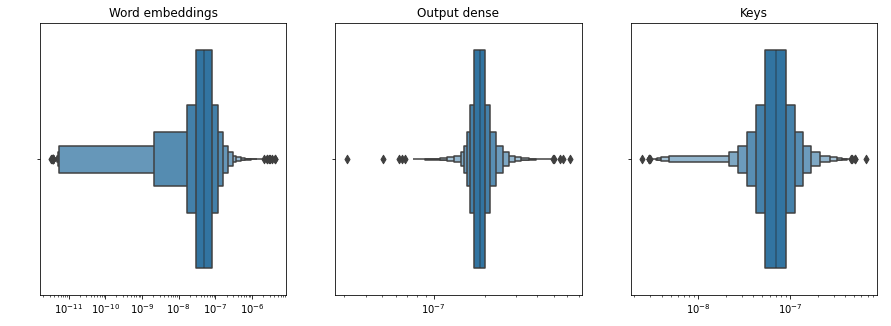}
    \caption{Row-wise Fisher Information distribution for three different modules in the BERT$_{BASE}$ model. The Fisher information values are passed as importance weights in the autoencoder loss function to help the compression model focus more on the module's important weights. \Cref{sec:appendix1:fisher} discusses different transformations applied to Fisher information to help the compression model handle outlier Fisher information values.}
    \label{fig:fisher_dist}
\end{figure}

We present the distributions of the row-wise Fisher Information for the upstream dataset (i.e., masked language modeling on the WikiText-103 dataset) in~\Cref{fig:fisher_dist}. We notice that each distribution contains some outliers which point to the potentially irrelevant weights in the considered weight matrix. In this section, we demonstrate that the weights' relevance information may be leveraged in the compression process to improve the quality of compressed modules.

\citet{hsu2022language} used the Fisher Information directly on the original model weights in their
Fisher-Weighted SVD (FWSVD) approach:

\begin{equation}
\label{eq:fwsvd}
W \approx FWSVD(W) = \hat{I}^{-1} SVD (\hat{I}W).
\end{equation}

In contrast, our method does not modify the original weight matrices but rather uses the Fisher Information in the loss function to help the model focus more on important weights. Moreover, we apply different transformations on the original Fisher Information values to modify the relative importance of the module weights to reduce the undesirable influence of outliers (\Cref{sec:appendix1:fisher} discusses various transformations we experimented with for different modules).

\input{tables/fisher_table_main_text.tex}

\paragraph{Results}
\Cref{tab:maintext_fisher_usage_comparison} presents the benefits of incorporating sensitivity for compression of token embeddings where both upstream perplexity and downstream task performance is improved. 
We further demonstrate the positive influence of Fisher information for three transformer modules in \Cref{tab:fisherusagecomparison} (in~\Cref{sec:appendix1:fisher}) for both autoencoder and SVD methods.
Additionally, \Cref{tab:fisher} (in~\Cref{sec:appendix1:fisher}) presents the compression performance provided by AE using different Fisher Information transformations. We observe that incorporating the Fisher Information with batch normalization into the compression process considerably improves the model perplexity as well as the downstream task performance.

\subsection{Comparison with other offline compression approaches}
\label{sec:abl:baselines}

In~\Cref{tab:fisherusagecomparison}, we compare our approach with the most popular matrix factorization method, namely Singular Value Decomposition~(SVD), for the compression of three different transformer modules. We may see that our approach outperforms or is competitive with SVD in most settings.
Additionally, in~\Cref{sec:abl:kronecker_tucker}, we discuss the poor performance of Kronecker Product and Tucker Decomposition (as two other factorization-based methods) in the offline compression setting. We also compare our solution to a non-factorization baseline, namely pruning, and show that our autoencoder-based method also outperforms it in most studied settings (see~\Cref{tab:pruning} in~\Cref{sec:abl:kronecker_tucker}).


\subsection{Compressing multiple types of transformer modules}
\label{sec:abl:finalres}

\input{tables/final_table.tex}

For the final experiment, we took into consideration all the analysis insides presented in this work and prepared the concluding experiment on offline compression methods. In this experiment, we compressed multiple BERT$_{BASE}$ weight matrices at the same time (token embedding matrix, all key matrices and all output-dense matrices). We compared the offline compression quality produced by the autoencoder approach and the baseline SVD factorization matrix method.

For the compression with SVD we classically compressed each matrix separately. We tested three different seeds and various number of iterations for SVD algorithm. For the autoencoder approach, we compressed considered matrices by selecting appropriate mechanisms based on our ablation study. For both SVD and our approach, we report the median of the final scores from the experiments with three different seeds to exclude potential outliers. The compressed matrices with the lowest perplexity were applied into BERT$_{BASE}$ that was then fine-tuned on various NLP tasks (MRPC, SST-2, RTE, QNLI and QQP). In the resulting table we reported the median of the final scores to exclude potential outliers. 

Final experiment results are presented in~\Cref{tab:fullbertcompression}. Our approach consistently outperforms the SVD baseline on all tested downstream tasks.\footnote{The hyperparameter setting for autoencoder is provided in~\Cref{tab:appendix:ae_hyperparameters} in ~\Cref{sec:appendix:ae_hyperparameters}}

\subsection{Compression time}
Generally, compressing modules using autoencoder and SVD takes a comparable amount of time.  However, using concatenated mode (as proposed in our paper) speeds up this process significantly. In~\Cref{sec:appendix:compressiontimes}, we report compression times for different modules (\Cref{tab:appendix:compressiontimes}) and the compressed models' inference and fine-tuning time.



%% file: tables/fisher_table_main_text.tex
\begin{table}[h]
    \small
    \centering
    \resizebox{\columnwidth}{!}{%
    \begin{tabular}{@{}ccrrlrrlrrl@{}}
        \toprule
        CR & Method & \multicolumn{1}{c}{Perplexity} & \multicolumn{1}{c}{\begin{tabular}[c]{@{}c@{}}SST-2\\ (Acc)\end{tabular}} & \multicolumn{1}{c}{\begin{tabular}[c]{@{}c@{}}MRPC\\ (F1/Acc)\end{tabular}} \\ \midrule
        \multirow{2}{*}{3} & AE & 118.41 & 91.97 & \multicolumn{1}{r}{\textbf{88.53 / 84.31}} \\
         & AE+Fisher & 33.27 & \textbf{92.55} & 88.36 / 83.33 \\ \midrule
        \multirow{2}{*}{10} & AE & 712.98 & 88.07 & \multicolumn{1}{r}{85.87 / 80.88} \\
         & AE+Fisher & 250.59 & \textbf{89.33} & \textbf{87.27 / 81.62}  \\ \midrule
        \multirow{2}{*}{25} & AE & 4926.08 & 82.80 & \multicolumn{1}{r}{84.19 / 77.45} \\
         & AE+Fisher & 2728.41 & \textbf{83.83} & \textbf{84.35 / 77.70} \\ \bottomrule
    \end{tabular}%
    }
    \caption{The effect of adding the Fisher Information to the autoencoder-based (AE) compression of token embeddings. We report the compressed BERT$_{BASE}$ upstream task perplexity (on the WikiText-103 dataset) and the downstream performance over two GLUE tasks. Each AE result represents a median from 3 runs with different seeds.}
    \label{tab:maintext_fisher_usage_comparison}
\end{table}

%% file: tables/final_table.tex
\begin{table*}[h!]
    \small
    \centering
    \vskip 0.1in
    \begin{tabular}{@{}ccrrlrrlrrl@{}}
         
         \midrule
         CR & Architecture & \makecell{MRPC \\ (F1/Acc)} & \makecell{SST-2 \\ (Acc)} & \makecell{RTE \\ (Acc)} & \makecell{QNLI \\ (Acc)} & \makecell{QQP \\ (F1/Acc)} \\
         \midrule
         
         1 & BERT$_{BASE}$ &  88.85/84.07 & 92.32 & 65.70	 & 90.66 & 87.49/90.71 \\ 
         \midrule
         
        \multirow{2}{*}{\begin{tabular}[x]{@{}c@{}}3\end{tabular}}
         & SVD & 84.35/77.45 & 86.70 & 62.09 & 85.61 & 84.64/88.38 \\
         & Our & \textbf{85.25/77.77} & \textbf{90.25} & \textbf{62.45} & \textbf{88.68} & \textbf{86.07/89.71}\\
         \midrule
         
         \multirow{2}{*}{\begin{tabular}[x]{@{}c@{}}10\end{tabular}}
         & SVD & 78.46/68.38 & 82.00 & 52.71 & 77.48 & 79.36/83.45\\
          & Our & \textbf{81.45/71.08} & \textbf{83.94} & \textbf{57.76} & \textbf{81.48} & \textbf{81.57/85.75}\\
         \midrule
         
         \multirow{2}{*}{\begin{tabular}[x]{@{}c@{}}25\end{tabular}}
         & SVD & 77.10/66.42 & 78.33 & 53.43 & 62.66  & 74.21/79.10\\
         & Our & \textbf{81.60/71.32} & \textbf{80.05} &  \textbf{55.23} & \textbf{72.96} & \textbf{78.45/83.11}\\
         \bottomrule
    \end{tabular}
    \caption{Final BERT$_{BASE}$ model compression (token embedding matrix, all key matrices, and all output-dense matrices). The baseline SVD algorithm compresses each matrix separately. Our autoencoder-based approach, incorporates mechanisms developed in the ablation study presented in this work (see~\Cref{tab:appendix:ae_hyperparameters} for the detailed AE design choices). For each setting, we present the median score from experiments with three different seeds. Our approach consistently outperforms the classical factorization method.}
    \label{tab:fullbertcompression}
\end{table*}

%% file: conclusions.tex
\section{Conclusions}
\label{sec:conclusions}
This work comprehensively studies various methods for the offline compression of transformer language models. We analyze various changes in the proposed architecture and its optimization function. We test different input modifications and evaluate the compressed language model performance in each scenario. By analyzing various compression settings, we show that our autoencoder-based approach outperforms classical matrix factorization on various NLP downstream tasks. Furthermore, we believe the techniques analyzed in this study might also be useful for low-cost compression of  different weight matrices unrelated to language models.

%% file: limitations.tex
\section*{Limitations}
\label{sec:limit}

A limitation of our approach that we may identify is the need to analyze each module type to determine the best mechanisms for its compression. Our module-specific findings could be reflected in corresponding modules in other language models, but this would require further investigation. 
Additionally, A (reasonably-sized) unsupervised corpus must also be used for computing the Fisher Information for the compression procedure, which is more computationally demanding than other offline approaches suggested in this study.



%% file: ethic_statement.tex

%% file: acknowledgement.tex
\section*{Acknowledgements}

The work of Klaudia Bałazy was supported by the National Centre of Science (Poland) Grant No. 2020/39/D/ST6/01332. Klaudia Bałazy is affiliated with Doctoral School of Exact and Natural Sciences at the Jagiellonian University. The research of Jacek Tabor was carried out within the research project "Bio-inspired artificial neural network" (grant no. POIR.04.04.00-00-14DE/18-00) within the Team-Net program of the Foundation for Polish Science co-financed by the European Union under the European Regional Development Fund. Artur Kasymov work was supported by the National Centre of Science (Poland) Grant No. 2019/33/B/ST6/00894.

%% file: apendix.tex
\appendix

\section{Experiments}
\label{sec:appendix1}

In this section, we describe the general assumptions for the experiments and the specific setting of the hyperparameters for each individual experiment.

We train the autoencoder model using gradient descent procedure and Adam optimizer~\cite{adam2014}. In most of our experiments, resulting compressed modules are inserted into a pre-trained language model and fine-tuned on two different downstream tasks from GLUE benchmark~\cite{wang-etal-2018-glue}, MRPC and SST2, with a default learning rate $\lambda = 2\cdot10^{-5}$ proposed by Hugging Face Transformers\footnote{https://github.com/huggingface/transformers}~\cite{wolf2019huggingface}.

\subsection{Cosine distance objective}
\label{sec:appendix1:cos}

In~\Cref{tab:coskey} and~\Cref{tab:cosout}, we report the effect adding the cosine distance component to the compression objective for the keys and output-dense matrices from the BERT$_{BASE}$ model. 

\paragraph{Experimental setup}
We study the effect of the cosine distance component in the loss function formulated in~\Cref{eq:l2cosloss} on the compression quality of self-attention keys matrices and the fully-connected output-dense weight matrices. We compress each of these matrices from each layer separately and then apply all the compressed matrices from certain type (keys or output-denses) to the transformer model to evaluate the compression quality. We inspect the effect of various ratios between Euclidean distance component~($L_2$) and the cosine distance component~($CD$) in~\Cref{eq:l2cosloss}, namely 1:0, 10:1, 1:1, 1:10 which corresponds to $\beta \in \{0.0, 0.0909, 0.5, 0.909\}$. For each model we test different learning rates $\lambda \in \{5\cdot10^{-3}, 10^{-3}, 5\cdot10^{-4}, 10^{-4}\}$.

\begin{table}[h!]
    \small
    \centering
    \vskip 0.1in
    \begin{tabular}{@{}cccccrrlrrl@{}}
         
         \midrule
         \makecell{Compression\\ratio}  & \makecell{Cosine \\coefficient} & \makecell{Initial \\perplexity} & \makecell{MRPC \\ (Acc)} & \makecell{SST-2 \\(Acc)} \\
         \midrule
         
          \multirow{4}{*}{\begin{tabular}[x]{@{}c@{}}3\end{tabular}}
         & 0.0 & 22.9 & 74.5 & 91.0 \\
         & 0.0909 & 22.2 & 76.2 & 92.1 \\
         & 0.5 & 25.5 & 77.2 & 92.1 \\
         & 0.9091 & 27.0 & 76.2 & 91.4 \\
         \midrule
         
         \multirow{4}{*}{\begin{tabular}[x]{@{}c@{}}10\end{tabular}}
         & 0.0 & 45.8 & 72.5 & 91.5 \\
         & 0.0909 & 32.5 & 72.0 & 90.7 \\
         & 0.5 & 34.4 & 70.3 & 91.5 \\
         & 0.9091 & 43.2 & 71.3 & 91.6 \\
         \midrule
         
         \multirow{4}{*}{\begin{tabular}[x]{@{}c@{}}25\end{tabular}}
         & 0.0 & 116.9 & 70.3 & 90.6 \\
         & 0.0909 & 167.6 & 69.6 & 90.2 \\
         & 0.5 & 115.3 & 69.1 & 90.9 \\
         & 0.9091 & 507.7 & 69.3 & 90.1 \\
         \bottomrule
    \end{tabular}
    \caption{The impact of adding a component responsible for preserving the direction in the compression of the key matrices from the self-attention block in BERT$_{BASE}$ model. We report the best (across different learning rates) initial model's perplexity after compression as well as it's performance on a downstream tasks for the different cosine coefficients ($\beta$ in \Cref{eq:l2cosloss}). Adding a component to the compression objective that aims to preserve the direction of the vectors in reconstructed matrix may bring a slight improvement on the quality of the result.}
    \label{tab:coskey}
\end{table}

\begin{table}[h!]
    \small
    \centering
    \vskip 0.1in
    \begin{tabular}{@{}cccccrrlrrl@{}}
         
         \midrule
         \makecell{Compression\\ratio}  & \makecell{Cosine \\coefficient} & \makecell{Initial \\perplexity} & \makecell{MRPC \\ (Acc)} & \makecell{SST-2 \\(Acc)} \\
         \midrule
         
          \multirow{4}{*}{\begin{tabular}[x]{@{}c@{}}3\end{tabular}}
         & 0.0 & 327.4 & 83.1 & 91.0 \\
         & 0.0909 & 332.8 & 82.3 & 91.5 \\
         & 0.5 & 599.2 & 79.4 & 90.5 \\
         & 0.9091 & 1510.4 & 75.2 & 90.5 \\
         \midrule
         
         \multirow{4}{*}{\begin{tabular}[x]{@{}c@{}}10\end{tabular}}
         & 0.0 & 1396.5 & 74.5 & 88.1 \\
         & 0.0909 & 1917.8 & 74.5 & 85.9 \\
         & 0.5 & 2099.3 & 72.0 & 84.9 \\
         & 0.9091 & 2326.1 & 71.8 & 84.0 \\
         \midrule
         
         \multirow{4}{*}{\begin{tabular}[x]{@{}c@{}}25\end{tabular}}
         & 0.0 & 1988.0 & 73.8 & 84.5 \\
         & 0.0909 & 2099.0 & 73.8 & 85.3 \\
         & 0.5 & 2144.7 & 72.5 & 84.7 \\
         & 0.9091 & 2148.6 & 72.5 & 84.6 \\
         \bottomrule
    \end{tabular}
    \caption{The impact of adding a component responsible for preserving the direction in the compression of the output-dense matrices in BERT$_{BASE}$ model. Different cosine coefficients refer to $\beta$ in \Cref{eq:l2cosloss}. We report the score for the model with the best initial perplexity across different learning rates as well as it's performance on MRPC and SST2 downstream tasks.}
    \label{tab:cosout}
\end{table}


\subsection{Concatenated and separated weight matrices}
\label{sec:appendix1:concsep}

\begin{figure*}[ht]
\centering
    \includegraphics[scale=0.3]{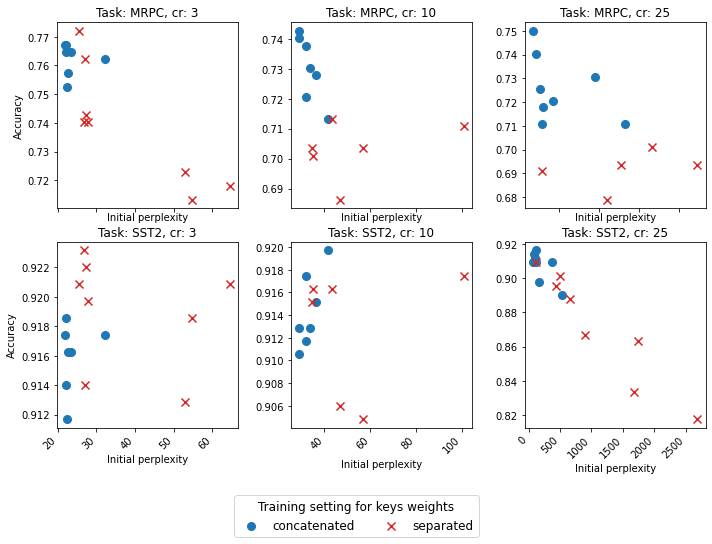} \hfill 
    \centering
    \includegraphics[scale=0.3]{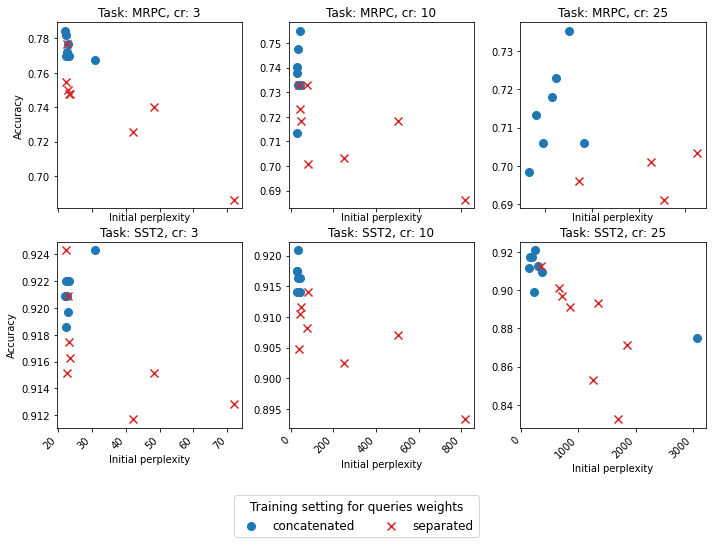}
    \includegraphics[scale=0.3]{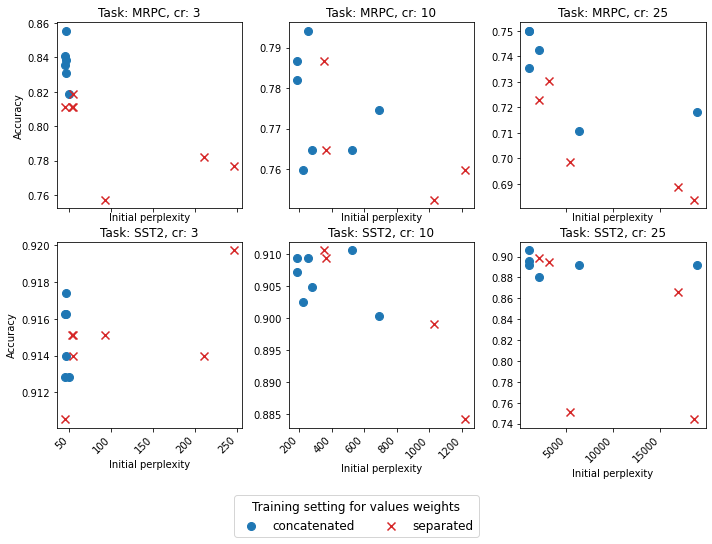}
    \caption{\label{fig:concvssep2}Initial perplexity and downstream tasks performance with separated and concatenated compression mode for output-dense weights matrices, queries weights matrices and values weights matrices in the BERT model.}
\end{figure*}

In~\Cref{fig:concvssep2} we present the performance achieved by models with compressed key, query and value matrices when using separated and concatenated mode in the compression process.

\paragraph{Experimental setup}
With the objective of comparing the compression of separate and concatenated matrices, we analyze the various matrices in the transformer architecture: the key, query, and value matrices from the self-attention module and output-dense matrix (one of the fully connected end matrices). We optimize loss function described previously in~\Cref{eq:l2cosloss} with 1:1 and 1:10 ratios for the $l_2$~norm loss coefficient and cosine distance coefficient, respectively. The decoder in the autoencoder model is a single fully connected layer. The model is trained with different learning rates $\lambda \in \{5\cdot10^{-3}, 10^{-3}, 5\cdot10^{-4}, 10^{-4}\}$.

\subsection{Initial perplexity vs downstream tasks performance}
\label{sec:appendix1:initperp}

\begin{figure*}[ht]
    \centering
    \includegraphics[scale=0.34]{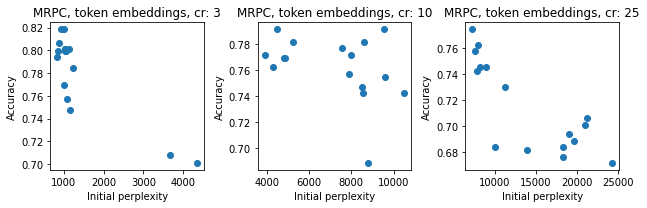}
    \includegraphics[scale=0.34]{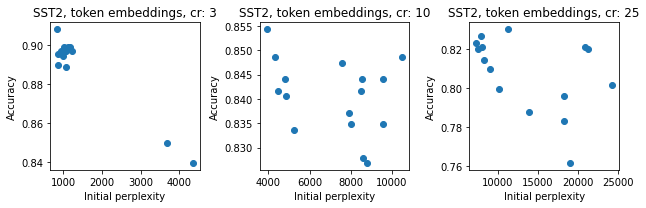}
    \includegraphics[scale=0.34]{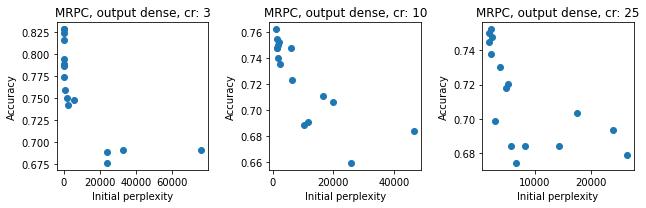}
    \includegraphics[scale=0.34]{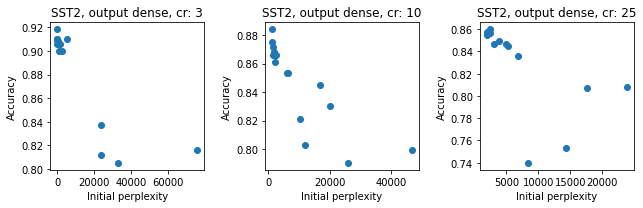}
    \includegraphics[scale=0.34]{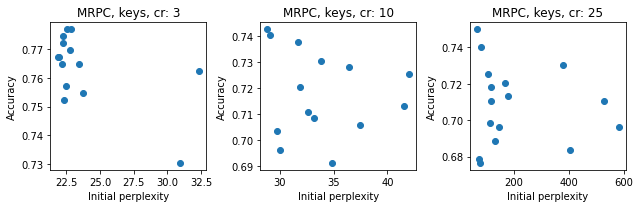}
    \includegraphics[scale=0.34]{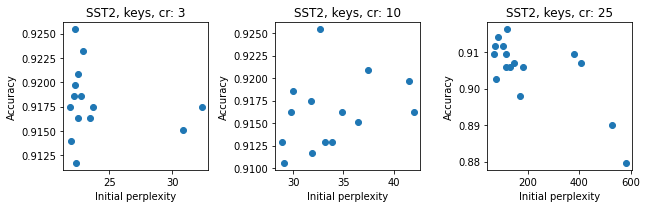}
    \includegraphics[scale=0.34]{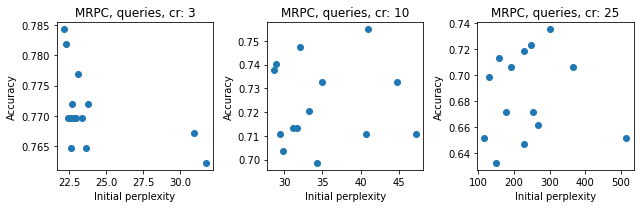}
    \includegraphics[scale=0.34]{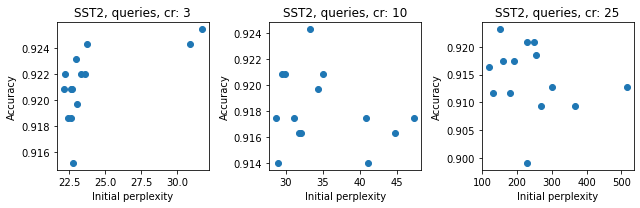}
    \includegraphics[scale=0.34]{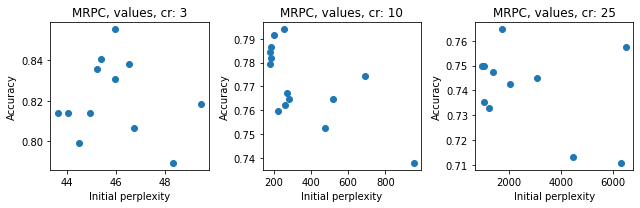}
    \includegraphics[scale=0.34]{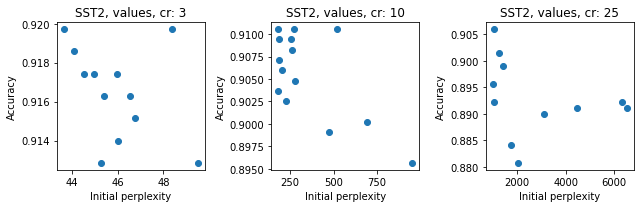}
    \caption{\label{fig:perplvsacc2}Initial perplexity and final accuracy on MRPC and SST2 downstream tasks for token embeddings, output-dense, keys, queries and values BERT$_{BASE}$ matrices compression.}
\end{figure*}

In~\Cref{fig:perplvsacc2} we present the initial perplexity and the performance on MRPC ans SST2 downstream tasks for the language model with various compressed modules.

\paragraph{Experimental setup}
We examine the initial masked language model perplexity and downstream tasks performance relation for token embedding matrix, self-attention matrices: keys, queries, values and final output-dense matrices. For simplicity of the experiment, we use the linear decoder in the autoencoder model. The loss and learning rate combinations for the models' training are the same as in experiment in previous section. We conducted all experiments with two different seeds to obtain more reliable correlation information.

\subsection{Linear and non-linear decoder module}
\label{sec:appendix_linear}

\begin{figure}[h]
    \centering
    \includegraphics[width=0.95\columnwidth]{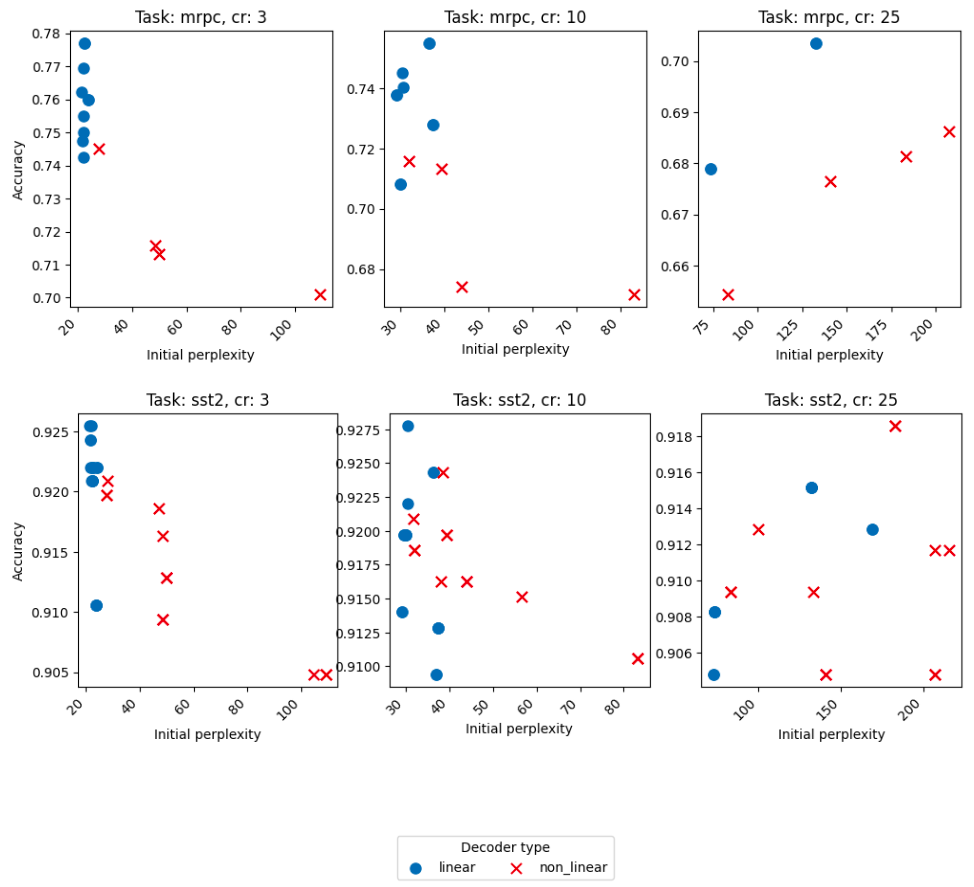}
    \caption{Initial perplexity and downstream tasks performance when using linear and non-linear decoder
module in the autoencoder model for the compression of the keys matrix. Using a linear
decoder generally appears to be a better choice for this matrix.}
    \label{fig:lin_non_lin_key}
\end{figure}

In~\Cref{fig:lin_non_lin_key} we present the initial perplexity and final downstream tasks performance achieved when using linear and non-linear decoder in the autoencoder model for key matrices compression. Unlike token embeddings for key matrices, the linear models outperform non-linear ones in most scenarios. Similar set of experiments on output-dense matrices also showed that linear models outperform non-linear ones.

\paragraph{Experimental setup}
We conduct experiments with the following settings: linear encoder/decoder and non-linear encoder/decoder. For non-linear encoder/decoder case, we examine architectures with 1 and 2 hidden layers. As non-linear activation functions we use LeakyReLU~\cite{leakyrelu} and Tanh. We investigate the loss configurations (~\Cref{eq:l2cosloss}) with 1:0, 1:1, 1:10 and 1:100 ratios of the $l_2$~norm loss to the cosine distance loss.


\subsection{Preserving vector norm}
\label{sec:app:preservingnorm}

\begin{figure}[h]
    \includegraphics[width=0.95\columnwidth]{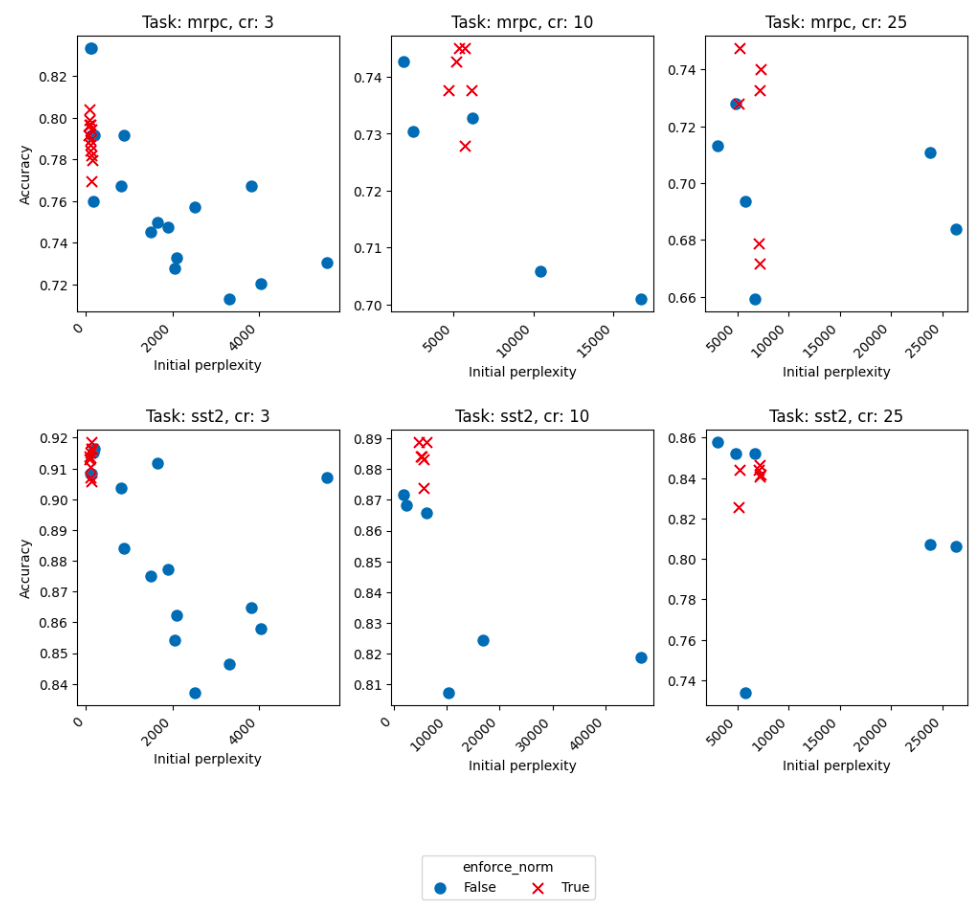}
    \caption{The effect of preserving $l_2$ vector norm on the perplexity and downstream tasks performance while compressing the output-dense matrices. We can see that enforcing norm for compression of this module generally improves the result.}
    \label{fig:enforce_norm_od}
\end{figure}

In~\Cref{fig:enforce_norm_od} we show the potential benefits of preserving $l_2$ vector norm during compression of the output-dense matrices for two different downstream tasks .

\paragraph{Experimental setup}
We repeat the same set of experiments as in the previous section~(\Cref{sec:abl:linnonlin}), but each experiment is executed with and without the preserving vector norm option enabled.


\subsection{Sensitivity}
\label{sec:appendix1:fisher}

In~\Cref{tab:fisherusagecomparison} and~\Cref{tab:fisher} we report the positive impact of adding Fisher Information (with different coefficient transformations) for weight matrix compression for both autoencoder and SVD approaches.

\paragraph{Experimental setup}

We precompute the Fisher Information coefficients for token embeddings, self-attentions keys and output-dense weight matrices. We apply them the best hyperparameter setting obtained from the previous experiments. We use various transformations for Fisher Information coefficients: exponential transformation $x^a$ with $a=0.1, 0.5, 0.9, 2.0$, logarithmic transformation $log_e(x)+C$ with a $C$ equals to minimum value so that all $x$ are positive, logarithmic transformation $log_e(x)+C+10$, and raw Fisher Information coefficients without any transformation (\emph{Vanilla}). For some transformations, we also add a batch sum normalization (\emph{+BN}). We also report results without using Fisher Information (\emph{No Fisher}).

\begin{table*}
\resizebox{\textwidth}{!}{%
\begin{tabular}{@{}ccrrlrrlrrl@{}}
\toprule
\multicolumn{1}{l}{} & \multicolumn{1}{l}{} & \multicolumn{3}{c}{Word embeddings} & \multicolumn{3}{c}{Output dense} & \multicolumn{3}{c}{Keys} \\ 

CR & Method & \multicolumn{1}{c}{Perplexity} & \multicolumn{1}{c}{\begin{tabular}[c]{@{}c@{}}SST-2\\ (Acc)\end{tabular}} & \multicolumn{1}{c}{\begin{tabular}[c]{@{}c@{}}MRPC\\ (F1/Acc)\end{tabular}} & \multicolumn{1}{c}{Perplexity} & \multicolumn{1}{c}{\begin{tabular}[c]{@{}c@{}}SST-2\\ (Acc)\end{tabular}} & \multicolumn{1}{c}{\begin{tabular}[c]{@{}c@{}}MRPC\\ (F1/Acc)\end{tabular}} & \multicolumn{1}{c}{Perplexity} & \multicolumn{1}{c}{\begin{tabular}[c]{@{}c@{}}SST-2\\ (Acc)\end{tabular}} & \multicolumn{1}{c}{\begin{tabular}[c]{@{}c@{}}MRPC\\ (F1/Acc)\end{tabular}} \\ \midrule

\multirow{4}{*}{3} & SVD & 1842.38 & 89.68 & 83.71 / 75.49 & 99.98 & \textbf{91.74} & \textbf{88.77 / 84.07} & 22.04 & 92.32 & 85.30 / 77.45 \\
 & SVD+Fisher & 20.20 & 91.86 & \textbf{88.62} / 83.82 & 71.05 & \textbf{91.74} & 88.74 / 83.82 & 20.67 & 91.74 & \textbf{86.22 / 78.92} \\
 & AE & 118.41 & 91.97 & \multicolumn{1}{r}{88.53 / \textbf{84.31}} & 70.69 & \textbf{91.74} & \multicolumn{1}{r}{86.17 / 78.92} & 22.35 & \textbf{92.43} & \multicolumn{1}{r}{84.48 / 76.23} \\
 & AE+Fisher & 33.27 & \textbf{92.55} & 88.36 / 83.33 & 64.67 & 91.40 & 85.58 / 77.94 & 22.27 & 92.32 & 84.44 / 75.98 \\ \midrule
\multirow{4}{*}{10} & SVD & 13196.30 & 83.37 & 82.85 / 74.02 & 989.18 & 89.56 & 83.92 / 75.49 & 30.75 & \textbf{91.97} & \textbf{81.54 / 73.04} \\
 & SVD+Fisher & 65.17 & 87.73 & 85.01 / 77.70 & 723.61 & \textbf{90.14} & \textbf{84.76 / 76.47} & 41.49 & 91.74 & 81.00 / 72.06 \\
 & AE & 712.98 & 88.07 & \multicolumn{1}{r}{85.87 / 80.88} & 1197.19 & 88.42 & \multicolumn{1}{r}{83.97 / 75.49} & 29.01 & 91.40 & \multicolumn{1}{r}{80.42 / 72.30} \\
 & AE+Fisher & 250.59 & \textbf{89.33} & \textbf{87.27 / 81.62} & 1249.62 & 89.45 & 84.64 / 75.98 & 29.30 & 91.40 & 81.34 / 72.79 \\ \midrule
\multirow{4}{*}{25} & SVD & 20178.74 & 77.64 & 82.33 / 71.81 & 1603.45 & \textbf{88.19} & 84.09 / 75.25 & 52.50 & 90.71 & 78.19 / 69.36 \\
 & SVD+Fisher & 913.23 & 75.23 & 82.02 / 73.77 & 1205.34 & 87.27 & 83.88 / 74.75 & 80.14 & 91.17 & 74.36 / 65.69 \\
 & AE & 4926.08 & 82.80 & \multicolumn{1}{r}{84.19 / 77.45} & 1462.41 & 85.61 & \multicolumn{1}{r}{82.02 / 72.06} & 69.24 & \textbf{91.74} & \multicolumn{1}{r}{\textbf{78.78 / 69.36}} \\
 & AE+Fisher & 2728.41 & \textbf{83.83} & \textbf{84.35 / 77.70} & 1453.44 & 87.56 & \textbf{84.54 / 75.98} & 73.28 & 90.83 & 78.40 / 69.61 \\ \bottomrule
\end{tabular}%
}
\caption{The effect of adding the Fisher Information to the SVD-based and autoencoder-based (AE) compression. We report the BERT$_{BASE}$ upstream task perplexity and the downstream tasks final scores. Each autoencoder result represents a median from 3 runs with different seeds and each SVD score is a result of the best iteration from run with one seed. For autoencoder model compression we selected the Fisher Information transformation for each of the compressed modules based on the results from~\Cref{tab:fisher}~($x^{0.5} + BN$ for word embeddings; $x^{2.0} + BN$ for output-dense matrices; $log_e(x) + C + 10$ for key matrices).}
\label{tab:fisherusagecomparison}
\end{table*}


\begin{table*}[htb]
\resizebox{\textwidth}{!}{%
\begin{tabular}{@{}lrrlrrlrrl@{}}
\toprule
\multicolumn{1}{c}{} & \multicolumn{3}{c}{Word embeddings} & \multicolumn{3}{c}{Output-dense} & \multicolumn{3}{c}{Keys} \\ 
\multicolumn{1}{c}{CR=3} & \multicolumn{1}{c}{Perplexity} & \multicolumn{1}{c}{\begin{tabular}[c]{@{}c@{}}SST-2\\ (Acc)\end{tabular}} & \multicolumn{1}{c}{\begin{tabular}[c]{@{}c@{}}MRPC\\ (F1/Acc)\end{tabular}} & \multicolumn{1}{c}{Perplexity} & \multicolumn{1}{c}{\begin{tabular}[c]{@{}c@{}}SST-2\\ (Acc)\end{tabular}} & \multicolumn{1}{c}{\begin{tabular}[c]{@{}c@{}}MRPC\\ (F1/Acc)\end{tabular}} & \multicolumn{1}{c}{Perplexity} & \multicolumn{1}{c}{\begin{tabular}[c]{@{}c@{}}SST-2\\ (Acc)\end{tabular}} & \multicolumn{1}{c}{\begin{tabular}[c]{@{}c@{}}MRPC\\ (F1/Acc)\end{tabular}} \\ \midrule
$x^{2.0}$ + BN & 30.88 & 92.32 & 87.97 / 82.84 & 64.67 & 91.40 & 85.58 / 77.94 & 22.44 & 92.20 & 83.94 / 75.25 \\
$x^{0.9}$ + BN & 24.97 & \textbf{92.55} & 88.32 / 83.09 & 62.93 & 91.51 & 86.13 / 78.92 & \textbf{22.18} & 92.20 & 84.04 / 75.49 \\
$x^{0.5}$ + BN & 33.27 & \textbf{92.55} & 88.36 / 83.33 & 65.07 & 91.40 & 85.81 / 78.19 & 22.32 & 91.97 & 83.89 / 75.25 \\
$x^{0.1}$ + BN & 58.08 & 91.86 & 86.47 / 81.13 & 68.19 & 91.40 & \textbf{ 86.17 / 78.92 }& 22.29 & 91.97 & 83.99 / 75.25 \\
Vanilla+BN & \textbf{24.68} & \textbf{92.55} & 87.25 / 82.11 & \textbf{62.60} & 91.63 & 86.08 / 78.43 & 22.39 & 92.20 & 83.70 / 74.51 \\
Vanilla & 564.37 & 87.96 & 87.32 / 82.35 & 66.91 & 90.77 & 85.99 / 78.68 & 22.73 & \textbf{92.55} & 83.62 / 74.75 \\
$log_e(x)+C+10$ & 68.46 & 92.43 & 86.91 / 81.62 & 68.61 & \textbf{91.86} & 86.04 / 78.68 & 22.27 & 92.32 & 84.44 / 75.98 \\
$log_e(x)+C$ & 236.24 & 92.09 & 86.75 / 81.13 & 68.17 & 91.74 & \textbf{86.17 / 78.92} & 22.26 & 91.97 & 84.36 / 75.74 \\ \midrule
No Fisher & 118.41 & 91.97 & \multicolumn{1}{r}{\textbf{88.53 / 84.31}} & 70.69 & 91.74 & \multicolumn{1}{r}{ \textbf{86.17 / 78.92}} & 22.35 & 92.43 & \multicolumn{1}{r}{\textbf{84.48 / 76.23}} \\ \bottomrule
\end{tabular}%
}
\resizebox{\textwidth}{!}{%
\begin{tabular}{lrrlrrlrrl}
\toprule
\multicolumn{1}{c}{} & \multicolumn{3}{c}{Word embeddings} & \multicolumn{3}{c}{Output-dense} & \multicolumn{3}{c}{Keys} \\
\multicolumn{1}{c}{CR=10} & \multicolumn{1}{c}{Perplexity} & \multicolumn{1}{c}{\begin{tabular}[c]{@{}c@{}}SST-2\\ (Acc)\end{tabular}} & \multicolumn{1}{c}{\begin{tabular}[c]{@{}c@{}}MRPC\\ (F1/Acc)\end{tabular}} & \multicolumn{1}{c}{Perplexity} & \multicolumn{1}{c}{\begin{tabular}[c]{@{}c@{}}SST-2\\ (Acc)\end{tabular}} & \multicolumn{1}{c}{\begin{tabular}[c]{@{}c@{}}MRPC\\ (F1/Acc)\end{tabular}} & \multicolumn{1}{c}{Perplexity} & \multicolumn{1}{c}{\begin{tabular}[c]{@{}c@{}}SST-2\\ (Acc)\end{tabular}} & \multicolumn{1}{c}{\begin{tabular}[c]{@{}c@{}}MRPC\\ (F1/Acc)\end{tabular}} \\ \midrule
$x^{2.0}$ + BN & 147.29 & 86.93 & 84.85 / 77.94 & 1249.62 & \textbf{89.45} & \textbf{84.64 / 75.98} & 30.78 & 91.40 & 79.73 / 70.10 \\
$x^{0.9}$ + BN & \textbf{95.32} & 88.30 & 85.22 / 78.92 & 1251.53 & 88.19 & 84.04 / 75.25 & 30.47 & 91.28 & 79.46 / 70.10 \\
$x^{0.5}$ + BN & 250.59 & \textbf{89.33} & \textbf{87.27 / 81.62} & 1195.41 & 88.53 & 83.81 / 75.00 & 29.81 & 91.40 & 80.60 / 71.57 \\
$x^{0.1}$ + BN & 567.88 & 87.16 & 84.73 / 77.21 & 1148.47 & 88.42 & 83.68 / 75.25 & 29.26 & 91.17 & 80.74 / 72.06 \\
Vanilla+BN & 136.84 & 88.99 & 85.86 / 79.90 & 1251.65 & 88.65 & 83.79 / 75.00 & 30.53 & 91.40 & 78.85 / 69.36 \\
Vanilla & 15994.88 & 79.70 & 81.22 / 68.38 & 1547.14 & 87.16 & 83.20 / 74.26 & 31.65 & \textbf{91.63} & 77.82 / 68.14 \\
$log_e(x)+C+10$ & 806.71 & 88.30 & 86.06 / 80.15 & \textbf{1145.25} & 88.42 & 83.71 / 75.00 & 29.30 & 91.40 & \textbf{81.34 / 72.79} \\
$log_e(x)+C$ & 854.19 & 88.07 & 86.25 / 80.39 & 1148.46& 88.99 & 84.09 / 75.25 & 29.25 & 91.17 & 80.14 / 72.06 \\ \hline
No Fisher & 712.98 & 88.07 & \multicolumn{1}{r}{85.87 / 80.88} & 1197.19 & 88.42 & \multicolumn{1}{r}{83.97 / 75.49} & \textbf{29.01} & 91.40 & \multicolumn{1}{r}{80.42 / 72.30}  \\ \bottomrule
\end{tabular}%
}

\resizebox{\textwidth}{!}{%
\begin{tabular}{@{}lrrlrrlrrl@{}}
\toprule
\multicolumn{1}{c}{} & \multicolumn{3}{c}{Word embeddings} & \multicolumn{3}{c}{Output-dense} & \multicolumn{3}{c}{Keys} \\ 
\multicolumn{1}{c}{CR=25} & \multicolumn{1}{c}{Perplexity} & \multicolumn{1}{c}{\begin{tabular}[c]{@{}c@{}}SST-2\\ (Acc)\end{tabular}} & \multicolumn{1}{c}{\begin{tabular}[c]{@{}c@{}}MRPC\\ (F1/Acc)\end{tabular}} & \multicolumn{1}{c}{Perplexity} & \multicolumn{1}{c}{\begin{tabular}[c]{@{}c@{}}SST-2\\ (Acc)\end{tabular}} & \multicolumn{1}{c}{\begin{tabular}[c]{@{}c@{}}MRPC\\ (F1/Acc)\end{tabular}} & \multicolumn{1}{c}{Perplexity} & \multicolumn{1}{c}{\begin{tabular}[c]{@{}c@{}}SST-2\\ (Acc)\end{tabular}} & \multicolumn{1}{c}{\begin{tabular}[c]{@{}c@{}}MRPC\\ (F1/Acc)\end{tabular}} \\ \midrule
$x^{2.0}$ + BN & 2779.96 & 80.73 & 83.99 / 75.98 & 1453.44 & 87.56 & \textbf{84.54 / 75.98} & 77.04 & 91.06 & 77.95 / 68.63 \\
$x^{0.9}$ + BN & \textbf{1983.19} & 83.49 & 82.08 / 73.04 & \textbf{1371.93} & \textbf{87.96} & 83.31 / 74.26 & 71.90 & 90.60 & 79.66 / 70.34 \\
$x^{0.5}$ + BN & 2728.41 & \textbf{83.83} & \textbf{84.35 / 77.70} & 1438.66 & 87.73 & 83.20 / 74.26 & 74.69 & 90.71 & \textbf{79.73 / 70.59} \\
$x^{0.1}$ + BN & 5502.34 & 82.45 & 84.12 / 76.96 & 1447.31 & 85.89 & 82.58 / 73.28 & 74.34 & 91.17 & 78.68 / 69.85 \\
Vanilla+BN & 2167.99 & 80.50 & 82.60 / 73.77 & 1384.83 & 87.10 & 83.87 / 75.49 & 71.37 & 90.71 & 78.97 / 69.85 \\
Vanilla & 14880.66 & 77.29 & 80.19 / 70.10 & 1748.90 & 86.35 & 82.26 / 73.28 & \textbf{66.53} & 91.17 & 76.92 / 67.65 \\
$log_e(x)+C+10$ & 5426.84 & 82.45 & 83.90 / 76.47 & 1449.46 & 86.93 & 82.89 / 73.53 & 73.28 & 90.83 & 78.40 / 69.61 \\
$log_e(x)+C$ & 4359.66 & 82.91 & 83.89 / 76.47 & 1446.55 & 86.58 & 81.79 / 72.06 & 76.81 & 91.06 & 77.78 / 68.63 \\ \midrule
No Fisher & 4926.08 & 82.80 & \multicolumn{1}{r}{84.19 / 77.45} & 1462.41 & 85.61 & \multicolumn{1}{r}{82.02 / 72.06} & 69.24 & \textbf{91.74} & \multicolumn{1}{r}{78.78 / 69.36} \\ \bottomrule
\end{tabular}%
}
\caption{Incorporating Fisher Information coefficients in the autoencoder-based compression process for different compression ratios (CR) on the SST2 and the MRPC tasks. The first column demonstrates the transformation(s) applied to Fisher information before being passed to autoencoder loss function (more details in~\Cref{sec:appendix1:fisher}).We also report the perplexity of the compressed model on the upstream task. Each score is the median of experiment results with three different seeds.}
\label{tab:fisher}
\end{table*}

\subsection{Comparison to other offline compression methods}
\label{sec:abl:kronecker_tucker}
\paragraph{Kronecker Product offline compression}
Inspired by a promising results achieved by using a Kronecker product for training the transformer model from scratch~\cite{shapeshifter} we have attempted to produce a compression of the original transformer matrices by using a Kronecker product of two matrices approximating the original matrix. We have trained these matrices using the gradient descent algorithm. Unfortunately, the results were unsatisfactory for each of the tested settings. For example, for concatenated key matrices and a compression ratio of 10 the perplexity for the Kronecker product was around 1500, while for the autoencoder perplexity below 50 is achieved in many different settings.

\paragraph{Tucker decomposition offline compression}
Moreover, we also experimented with the Tucker decomposition~\cite{de2000multilinear} as an offline compression method. For token embeddings compression, we observed that the compressed language model starts having high perplexities even in low compression ratios. For instance, for CR=3, the model perplexity becomes almost 1500, while the autoencoder model can achieve perplexities below 40 for the same compression ratio. This finding is consistent with the observation of \citep{ren2022exploring} that even randomly initializing the factorized tensors perform very close to the models initialized by e.g., tucker decomposition. Therefore, we also do not find Tucker decomposition an efficient method in the context of offline compression.

\paragraph{Pruning}
We also compare our proposed autoencoder framework with a pruning baseline as another offline compression baseline. The pruning algorithm here is based on PyTorch unstructured L1 pruning~\cite{paszke2019pytorch}. For this experiment, we compress token embedding, keys, and output-dense matrices using either autoencoder or pruning approaches. The models are evaluated on four GLUE tasks as presented in~\Cref{tab:pruning}. We can see that autoencoder-based compression outperforms pruning baseline in most studies settings, especially when higher compression ratios are studied.
\begin{table*}[h!]
    \small
    \centering
    \vskip 0.1in
    \begin{tabular}{@{}ccrrlrrlrrl@{}}
         
         \midrule
         CR & Architecture & \makecell{MRPC \\ (F1/Acc)} & \makecell{SST-2 \\ (Acc)} & \makecell{RTE \\ (Acc)} & \makecell{QNLI \\ (Acc)}\\
         \midrule
         
         1 & BERT$_{BASE}$ &  88.85/84.07 & 92.32 & 65.70	 & 90.66 \\ 
         \midrule
         
        \multirow{2}{*}{\begin{tabular}[x]{@{}c@{}}3\end{tabular}}
         & Pruning & \textbf{86.25/80.15} & \textbf{90.37} & 56.47 & \textbf{88.68}\\
         & Our & 85.25/77.77 & 90.25 & \textbf{62.45} & \textbf{88.68}\\
         \midrule
         
         \multirow{2}{*}{\begin{tabular}[x]{@{}c@{}}10\end{tabular}}
         & Pruning & 79.61/69.61 & 79.93 & 50.21 & 70.65\\
          & Our & \textbf{81.45/71.08} & \textbf{83.94} & \textbf{57.76} & \textbf{81.48} \\
         \midrule
         
         \multirow{2}{*}{\begin{tabular}[x]{@{}c@{}}25\end{tabular}}
         & Pruning & 80.86/69.36 & 75.11 & 51.97 & 60.57\\
         & Our & \textbf{81.6/71.32} & \textbf{80.05} &  \textbf{55.23} & \textbf{72.96} \\
         \bottomrule
    \end{tabular}
    \caption{Comparison of BERT$_{BASE}$ model compression using autoencoder (AE) and pruning approaches (compressing token embedding matrix, all key matrices and all output-dense matrices). For AE we present the median score from experiments with 3 different seeds.}
    \label{tab:pruning}
\end{table*}

\subsection{Hyperparameter setting for the final experiment}
This section presents the hyperparameter setting used for the~\Cref{sec:abl:finalres} where multiple modules (token embedding, keys and output-dense) are compressed together. The autoencoder hyperparameters setting for this experiment can be found in~\Cref{tab:appendix:ae_hyperparameters}.
\label{sec:appendix:ae_hyperparameters}

\begin{table*}[h!]
    \small
    \centering
    \vskip 0.1in
    \begin{tabular}{@{}ccccccclrrl@{}}
         
         \midrule
         Module & CR & \makecell{Learning \\ rate}  & \makecell{Cosine:L$_2$ \\ coefficients} & \makecell{Decoder} & \makecell{Enforce \\ norm} & \makecell{Fisher \\ transformation}\\
         \midrule
         
        \multirow{3}{*}{\begin{tabular}[x]{@{}c@{}}Token embeddings\end{tabular}}
         & 3 & $5\cdot10^{-4}$ & 10:1 & Non-linear (1 hidden layer) & Yes & $x^{0.5} + BN$ \\
         & 10 & $5\cdot10^{-4}$ & 10:1 & Non-linear (2 hidden layers) & Yes & $x^{0.5} + BN$ \\
         & 25 & $10^{-3}$ & 10:1 & Non-linear (1 hidden layer) & Yes & $x^{0.5} + BN$ \\
         \midrule

         \multirow{3}{*}{\begin{tabular}[x]{@{}c@{}}Output-denses\end{tabular}}
         & 3 & $10^{-3}$ & 1:10 & Linear & Yes & $x^{2.0} + BN$ \\
         & 10 & $10^{-4}$ & 1:10 & Linear & No & $x^{2.0} + BN$ \\
         & 25 & $5\cdot10^{-4}$ & 0:1 & Linear & No & $x^{2.0} + BN$ \\
         \midrule

          \multirow{3}{*}{\begin{tabular}[x]{@{}c@{}}Keys\end{tabular}}
         & 3 & $5\cdot10^{-4}$ & 0:1 & Linear & Yes & $log_e(x)+C+10$ \\
         & 10 & $5\cdot10^{-4}$ & 1:1 & Linear & No & $log_e(x)+C+10$ \\
         & 25 & $10^{-3}$ & 0:1 & Linear & No & $log_e(x)+C+10$ \\

         \bottomrule
    \end{tabular}
    \caption{The best hyperparameters for the BERT$_{BASE}$ model compression described in~\Cref{sec:abl:finalres}. In this experiment, token embedding matrix, all keys and all output-dense matrices are compressed using our proposed autoencoder-based framework.}
    \label{tab:appendix:ae_hyperparameters}
\end{table*}

\subsection{Compression time}
\label{sec:appendix:compressiontimes}

In~\Cref{tab:appendix:compressiontimes} we report compression times for keys, output-denses and token embeddings matrices from BERT$_{BASE}$ model using autoencoder and SVD compression approaches. For autoencoder we present times for both separated and concatenated matrices compression (as proposed in our paper) showing the advantage of using the latter approach. 

We additionally performed experiments to compare the inference time of uncompressed BERT$_{BASE}$ with the extreme case of CR=25 using our approach (AE) and the SVD baseline. Compressed modules are token embeddings, key, and output-dense modules. We observe that inference times are very similar, with a slight increase in inference time when our model is used. In particular, the evaluation times (in seconds) for BERT/SVD/AE were respectively 0.474/0.477/0.493 (for the MRPC dataset) and 0.987/0.997/1.034 (for the SST2 dataset). Moreover, fine-tuning the BERT$_{BASE}$ model using compressed modules from autoencoder for the bigger datasets in GLUE, namely QNLI and QQP, takes at most 15\% and 25\% longer than the SVD baseline, respectively. It is also worth noting that when a linear autoencoder is incorporated, the inference time is the same as the SVD baseline.

\begin{table*}[h!]
    \small
    \centering
    \vskip 0.1in
    \begin{tabular}{@{}ccccccc}

         \midrule
         Method & Mode & CR & \makecell{Token \\ embeddings} & Keys & Output-denses\\
         \midrule
         
         SVD & separated &  3 & $\sim$9.5min & ($\sim$5.5*12)min & ($\sim$6.0*12)min \\ 

         SVD & separated &  10 & $\sim$8.0min & ($\sim$5.5*12)min & ($\sim$6.0*12)min \\ 

         SVD & separated &  25 & $\sim$7.5min & ($\sim$5.5*12)min & ($\sim$6.0*12)min \\ 

         AE & separated/concatenated &  3 & $\sim$7.7min & ($\sim$5.5*12)min/$\sim$6.1min & ($\sim$6.0*12)min/$\sim$6.5min \\

         AE & separated/concatenated &  10 & $\sim$7.5min & ($\sim$5.5*12)min/$\sim$6.0min & ($\sim$6.0*12)min/$\sim$6.1min \\

         AE & separated/concatenated &  25 & $\sim$7.5min & ($\sim$5.5*12)min/$\sim$5.9min & ($\sim$6.0*12)min/$\sim$6.1min \\        
         \bottomrule
    \end{tabular}
    \caption{Training time to retrieve compressed modules (key, output-dense, and token embeddings) of BERT$_{BASE}$ model using autoencoder (AE) and SVD approach. For the AE, we provide training times for the separated and concatenated modes to demonstrate another benefit of using the concatenated version, given its much better training time.}
    \label{tab:appendix:compressiontimes}
\end{table*}